\documentclass[remotesensing]{mdpi} 
\usepackage[export]{adjustbox}
\usepackage{amsmath,amssymb}
\usepackage{graphicx}
\DeclareMathOperator{\E}{\mathbb{E}}
\usepackage{placeins}
\preto{\abstractkeywords}{\nolinenumbers}
\Title{Image fusion using symmetric skip autoencoder via an Adversarial Regulariser}

\Author{Snigdha Bhagat$^{1},\ddagger$, S.D.Joshi$^{2},\ddagger$ and Brejesh Lall$^{3},\ddagger$ }

\AuthorNames{Firstname Lastname, Firstname Lastname and Firstname Lastname}

\address{%
$^{1}$ \quad Affiliation 1; snigdha.bhagat@ee.iitd.ac.in\\
$^{2}$ \quad Affiliation 2; sdjoshi@ee.iitd.ac.in \\
$^{3}$ \quad Affiliation 3; brejesh@ee.iitd.ac.in }
\secondnote{These authors contributed equally to this work.} 

\abstract{It is a challenging task to extract the best of both worlds by combining the spatial characteristics of a visible image and the spectral content of an infrared image. In this work, we propose a spatially constrained adversarial autoencoder that extracts deep features from the infrared and visible images to obtain a more exhaustive and global representation. In this paper, we propose a residual autoencoder architecture, regularised by a residual adversarial network, to generate a more realistic fused image. The residual module serves as primary building for the encoder, decoder and adversarial network, as an add on the symmetric skip connections perform the functionality of embedding the spatial characteristics directly from the initial layers of encoder structure to the decoder part of the network. The spectral information in the infrared image is incorporated by adding the feature maps over several layers in the encoder part of the fusion structure, which makes inference on both the visual and infrared images separately. In order to efficiently optimize the network’s parameters, we propose an adversarial regulariser network which would perform supervised learning on the fused image and the original visual image since the visual image contains most of the structural content in comparison to the infrared image. In this paper we have thus focused on retaining the characteristics of the visual image perfectly and adding additional details from its infrared counterpart. }

\begin{document}
\section{Introduction} 
Information fusion is a technique to integrate relevant information from disparate sensors to merge, collate or juxtapose data in order to obtain a robust image output which can facilitate several subsequent processing tasks \cite{ref_article1}. When we are dealing with visible and infrared image fusion, the underlying issue is the limitation of bandwidth in capturing image data. The effectiveness of fusion is a very subjective process, and thus it is often assessed by the level of artifacts and abnormalities in the fused outcome. Since image fusion essentially involves generating new data from the distributions of multiple input images in the past few years, generative modelling techniques have been employed for the task of multi-modal image fusion \cite{ref_article20}\cite{ref_article21}\cite{ref_article22}. Visible images provide texture content and the details of the underlying structure. They are in accordance with the human visual system while infrared images capture the image content in a different frequency band. The infrared imaging, also referred to as thermal imaging, captures signals having wavelength higher than that of visible light, which are not visible to the human eye. It creates images based on differences in surface temperature by detecting infrared radiation (heat) that emerges from objects and their surrounding environment and is thus often used to improve the night time vision. The central motive of image fusion is to extract salient information from both the modalities and remove unnecessary details without creating artifacts in the fused image.  Image fusion thus plays a crucial role in video surveillance, modern military and satellite cloud imaging applications. Visible and infrared images thus form a pair of complementary data. In this paper we have employed an encoder-decoder architecture with residual connections regularised by an adversarial network; the encoder part has a downsample path that maps the input signal to a lower-dimensional space called the latent space from which the data is reconstructed by the decoder part which in turn provides an upsample path to transform the latent space to the original signal space. The network is optimised to obtain an efficient latent representation at the output of encoder such that the original signal can be reconstructed effectively. The residual connection, on the other hand, helps avoid the problem of model degradation caused due to increased depth of the network. The adversarial regulariser helps the autoencoder fusion network to generate a more realistic fused image that can effectively incorporate the characteristics of the visual image.
\section{Literature review}
Traditionally any image fusion problem can be formulated as:
\begin{equation}
    min [f_1(I,F)+f_2(V,F)]
\end{equation}
In the first term $f_1$ can be visualised as a map from infrared image \textbf{I} and the fused image \textbf{F} which is used to maximise the spectral relationship and in the second term $f_2$ is a map from the visual image \textbf{V} to the fused image that would in turn maximise the spatial relationship and generate a spectrally and spatially enhanced fused image. In the literature the image fusion problem has been addressed utilising different schemes including multi-scale transform \cite{ref_article1}\cite{ref_article2}\cite{ref_article3}, sparse representation \cite{ref_article4}\cite{ref_article5}, neural network \cite{ref_article6}\cite{ref_article7}, subspace \cite{ref_article8}\cite{ref_article9}, and saliency-based \cite{ref_article10}\cite{ref_article11} methods, hybrid model based approaches \cite{ref_article12}\cite{ref_article13}, and other methods\cite{ref_article14}\cite{ref_article15}. Any fusion framework involves three basic components: First, an image transformation model, second activity level measurement and third the formulation of fusion rule. Several deep learning framework based algorithms have also been successfully applied to find a solution to image fusion problem, due to their ability to extract image features. Yang et al.\cite{ref_article16} in their paper for fusion of multi-spectral(MS) and panchromatic images(PAN) have proposed a deep network based solution called the PanNet which specifically tries to embed domain based knowledge into the architecture by focusing on spectral and spatial information preservation of the fused output. In order to satisfy the spatial constraint, the ResNet is trained on high-frequency details of MS and PAN images to produce a higher resolution fused image and the spectral constraint is satisfied by combining the output of the spatial preserving network with the upsampled MS images. Since the fused output mainly depends on the output of the spatial preserving network, the spectral content fusion is a function of the spatial information; thus, it lacks stability and robustness.  Jinjiang Li et al.\cite{ref_article17} in his paper for multi-focus image fusion has executed the task of image fusion by decomposing the input images into low and high-frequency components since both of them carry unique information. Further two separate deep neural networks have been employed to train the high frequency and low-frequency sub-bands of the image separately instead of directly using the source image for end-to-end training. To generate fused low-frequency sub-band image, a siamese network is deployed to find high-level feature maps which are then used to find a fusion map for pixel-level fusion. For high-frequency sub-band fusion, a residual neural network is trained using a texture preserving loss function. On similar lines, Tang et al. \cite{ref_article18} improved the algorithm by proposing a pixel level CNN that would classify focused and the defocused pixels. These methods have successfully achieved robust and perceptually relevant state-of-the-art performance. Unlike multi-focus image fusion, visible and infrared images do not have a clear cut evaluation metric since the output can only be evaluated based on perceptual quality and ground truths cannot be constructed artificially. Besides, there is a dire shortage of publicly available databases for training deep networks. Despite all these exploring, there is no doubt that visible and infrared image fusion based on convolution neural networks is worth researching since deep neural networks can model complex linear and non-linear characteristics of the input images considerably. Jingchun Piao et al. \cite{ref_article19} in his paper has proposed a deep network for the fusion of infrared and visible images on different scales by using multi-scale wavelet decomposition. In order to determine the fusion rule, the Siamese network is used that determines the saliency of each pixel from the two images. Hui Li et al. \cite{ref_article20} in his paper has proposed visual and infrared image fusion by first decomposing the source images into base and detail part out of which the base parts are fused using weighted averaging. The detail part is fused using VGG network by extracting features at multiple layers in order to generate a weight map for fusion. Several versions of fused detail content are generated using the multi-layer features by employing the $l_1$ norm and weighted-average strategy. The final fused detail content was obtained by employing the max selection rule. The fused image is then constructed by a weighted combination of the base and detail content. The task of feature extraction is carried out using a pre-trained network, and the formulation of activity level measurement and fusion rule requires manual intervention, thus making the process highly unreliable. Since such architectures fail to provide an end-to-end solution without any manual intrusion. FusionGAN proposed by Jiayi Ma et al. \cite{ref_article21} tried to solve this problem by proposing an end-to-end architecture based on deep generative networks. FusionGAN posed this problem as an adversarial game where the generator performs the task of retaining the infrared thermal radiation information while incorporating the gradient information from the visual image and on the other hand discriminator tries to drive the fused image closer to a visible image.This has been proposed as a mini-max problem between the generator and discriminator. Ruichao Hou et al. \cite{ref_article22} also tried to overcome the same limitation in his paper and developed an dynamically adaptive end-to-end deep fusion framework called the visible and infrared image fusion network (VIF-Net), the deep network has been trained on a composite loss function that consists of M-SSIM that helps improve the perceptual quality of the image and total variation (TV) which helps improve the spatial quality of the image.

\section{Autoencoder constrained by Adversarial Regulariser}
Generative modelling is a domain of machine learning in which the network tries to learn the underlying distribution of the data from the given set of data points. Considering the data samples of a class as the training set the network tries to generate the best fit continuous distribution which, when sampled, can create new data points of the same class with some variations. But since it is not always possible to learn the exact distribution from the given data points so we try to model a distribution that can best approximate the true data. This is where neural networks come handy since they learn a function that can model the data distribution. Variational Autoencoders(VAE) and Generative adversarial networks (GAN) are two most commonly used approaches when it comes to generative modelling. Autoencoders serve as a useful tool when we need to obtain a compressed representation such that the data point can be perfectly reconstructed from such compressed latent space representation. In this paper, we propose the use of autoencoders for interpolation by a convex combination of the latent codes, which would, in turn, semantically mix the characteristics of both the input images. Since intuitively the distribution of fused image would be a weighted combination of the distribution of visible and infrared images.
\newline
In this paper, we propose to use adversarial training such that the autoencoder serves as the generator model, and the discriminator can be used as a classifier that would differentiate between the fused image and input images. Depending on the output, the error can be used to train the generator and the discriminator network. Adversarial training would help the generator to produce a more realistic fused image until and unless it can fool the discriminator model.

\section{Residual Networks}
\begin{figure}[t!]
\centering
\includegraphics[width=0.5\textwidth]{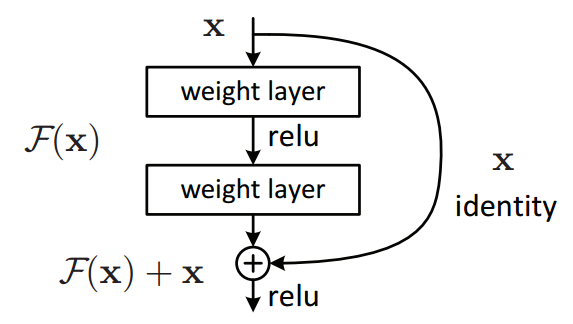}
\caption{Single Residual Block}
\end{figure}
Image classification has advanced in the past few years due to the availability of large datasets for training and powerful GPUs that has enabled the training of very deep architectures. Simonyan et al. \cite{ref_article26}, authors of VGG, successfully proved that accuracy can be increased by adding more and more layers to a network. Before this, in 2009, Yoshua Bengio \cite{ref_article27} gave convincing theoretical and mathematical evidence for the effectiveness of deeper neural network over their shallow counterparts. The residual network was first proposed by He et al. in \cite{ref_article24}, it was observed that as the depth of the network increases the accuracy of the network first saturates and then starts decreasing. This had nothing to do with overfitting, and thus the dropout layer could not solve this problem. It could be argued that this could be posed as an optimization problem since as the depth increases it becomes harder to train and propagate the error throughout the entire deep network due to vanishing gradients since the gradient is multiplied by the weight matrix at each step during back-propagation, thus if gradients are small due to successive multiplications its value would diminish. Although neural networks are universal function approximators, therefore, any deep or shallow neural network should be able to learn any simple or complex functions, but due to the curse of dimensionality and vanishing gradients, deep networks often are not able to learn simple identity mapping. Traditionally in a neural network, the output of one layer feeds the next layer, and in a residual network, the output would feed the next layer and another layer after 2-3 hops. To solve the issue of declining accuracy with increasing depth, the residual networks came in handy since they could learn simple mapping functions. As pointed out by Veit et al. \cite{ref_article25} this strategy also solves the problem of vanishing gradients since the error can be propagated efficiently to the initial few layers and they can also learn as fast as the last few layers. He has successfully visualized a deep residual network as an ensemble of several shallow networks with variable lengths. Thus the residual network becomes easy to optimize and can enjoy accuracy gains from significantly increased depth. ResNet can have a very deep network of up to 152 layers and still learn functions with a good accuracy since it has to learn the residual representation function instead of the signal representation.

\section{Proposed Method}
\subsection{Problem Formulation}
\begin{figure}[t!]
    \centering
    \includegraphics[scale = 0.7]{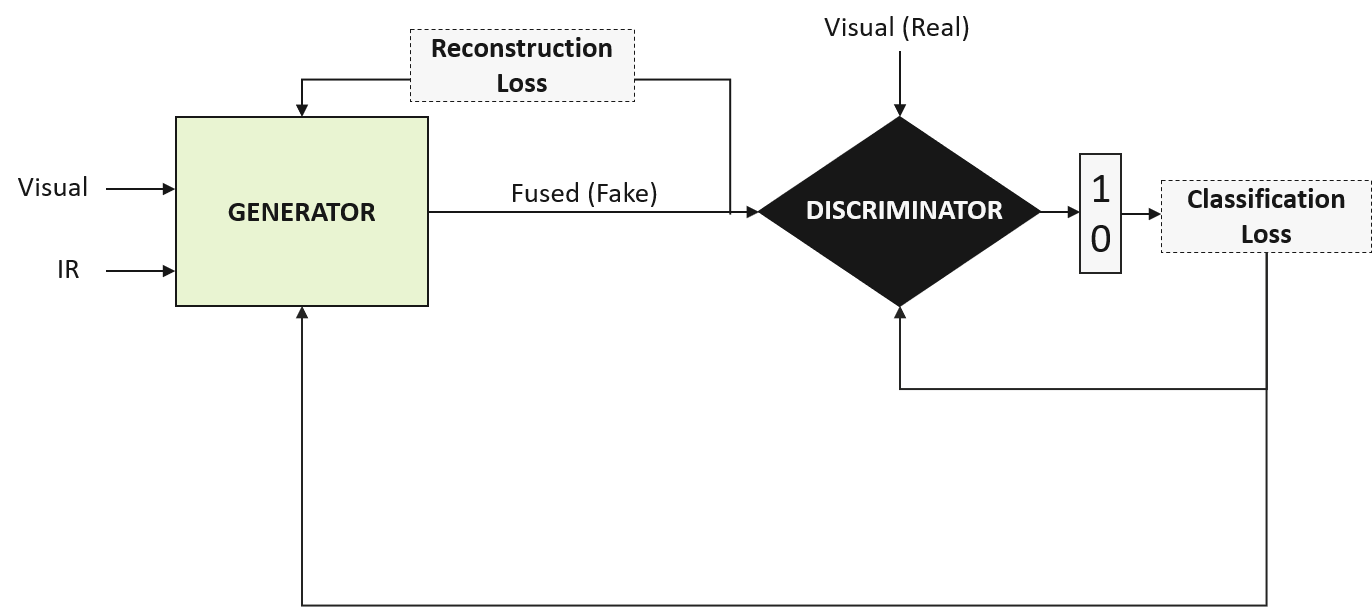}
    \caption{Overall Structure}
\end{figure}
In this paper, we propose a residual encoder-decoder architecture for image fusion along with a discriminator network that can perform the task of adversarial training, as shown in Figure 2. The architecture of the proposed modules have been shown in Figure 3 and Figure 4. It has three primary units ,namely, encoder, decoder and the discriminator block. The generator network is the residual encoder-decoder network, which tries to fuse the features of both the modalities by combining the latent space representations at several intermediate layers, as shown in Figure 3. Intuitively the feature maps obtained at several layers are fused in order to combine the content of both the images and since an autoencoder network tries to reconstruct data back from the reduced encoded representation, also called the latent space representation, in the most efficient manner the output of the decoder is the fused image. Thus the fused image (\textbf{F}) can be represented as RAE(V,I) where RAE is the residual autoencoder function. The generator network is conditioned by the reconstruction loss and a total variation loss which tries to maximise the texture content in the fused image as given in equation 4. The discriminator network is a classifier network that generates a scalar value which estimates the probability that a given image is a real image and not the output of the generator network. Thus the classification loss due to the discriminator network is also propagated to the generator network so that it can generate a more realistic image such that the discriminator would classify it as the input visual image rather than its fused counterpart. The discriminator is fed with only visual image and not the infrared image since most of the texture content is in the visual image the infrared image provides details which are mostly highlighted by huge contrast variations that are incorporated in the fused image due to the generator cost function. Since the objective is to gain the ability to distinguish between the real and generated samples cross entropy loss function has been employed in order to train the generator module. The overall training target of the generator module is to minimise the following objective function as given in equation 2, and the discriminator module also denoted as \textbf{D} would, in turn, try to maximise the same. 
\begin{equation}
    min_{RAE}max_{D}\{\E[\log(1 - D(F)]+ \E[\log(D(V)]\}
\end{equation}
The generator cost function is thus a combination of the reconstruction loss function also denoted as $L_{content}$ and the classification loss of the discriminator module as denoted as $L_{gen|disc}$ which serves as a regulariser as given in Equation 3. The discriminator, on the other hand, has to be trained to distinguish between real and fake data is trained to minimise the cross-entropy loss as can be seen from Equation 5, since cross-entropy can quantify the difference between two probability distributions for a given random variable or set of random variables. $\alpha$ and $\beta$ are hyper-parameters whose values can be varied in accordance with the requirement. In our simulations, we have tuned the value of hyper-parameters using the grid search method.
\begin{equation}
    L_{generator} = L_{content} + L_{gen|disc}
\end{equation}
\begin{equation}
    L_{content} = \beta \E[{(F-I)^{2}}]+(1-\beta) \E[{(F-V)^{2}}] + \alpha \left\|(F-V)\right\|_{TV} 
\end{equation}
\begin{equation}
    L_{disc} = -\E[\log(1 - D(F))]-\E[\log(D(V))]
\end{equation}
\begin{equation}
    L_{gen|disc} = \E[\log(1 - D(F))]
\end{equation}

\begin{figure}
    \centering
    \includegraphics{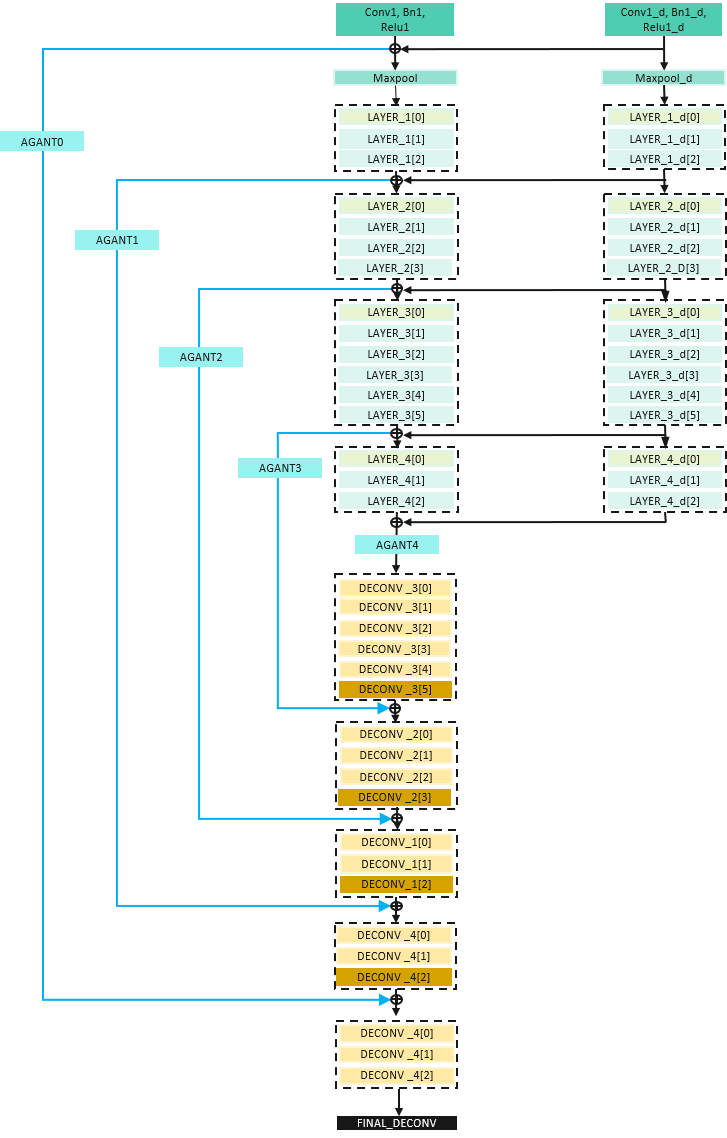}
    \caption{Fusion Autoencoder Module(Generator)}
\end{figure}

\begin{figure}
    \centering
    \includegraphics[scale = 1.3 ]{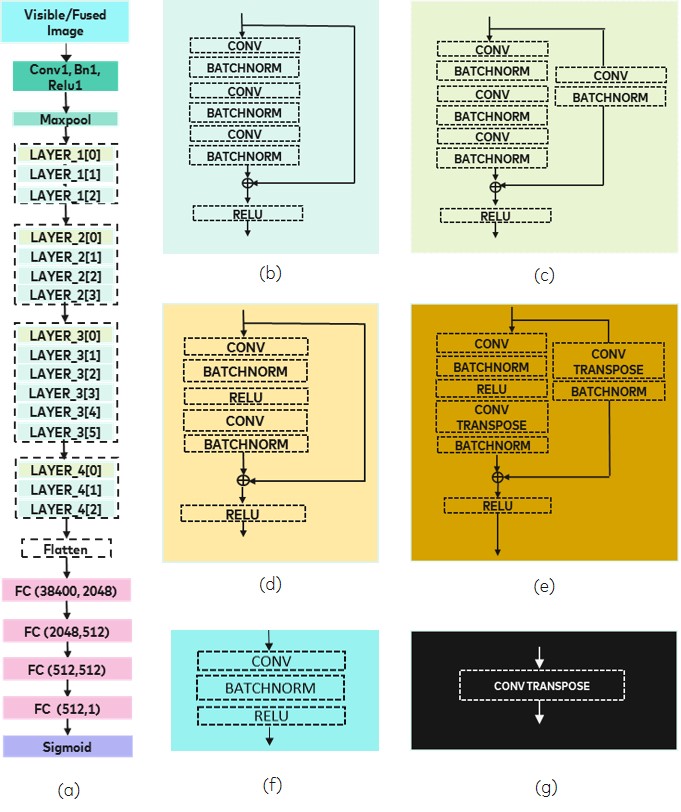}    
    \caption{(a)Network Architecture of Discriminator Module (b)Bottleneck Block without downsampling operation (c)Bottleneck Block with downsampling operation (d) Transbasic Block without upsampling operation (e) Transbasic Block with upsampling operation (f) Agant Layer (g) Final Deconv Layer}
\end{figure}

\subsection{Network Architecture}
The architecture of the generator module is shown in figure 3. The network architecture of the encoder network has been adopted from the ResNet architecture proposed by Kaiming He et al. \cite{ref_article24}. The idea of combining the latent space representation in order to fuse the content of two images was inspired from works of David Berthelot et al. \cite{ref_article29} on interpolation of data by fusing the latent space representation of images and Jindong Jiang et al. \cite{ref_article30} on semantic segmentation in which RGB image and the depth image are combined in order to segment different objects in a room.
The generator architecture is a residual autoencoder network with symmetric skip connections. Layers 1 - 4 constitutes the encoder part of the network and Layer 5-10 is the decoder part of the architecture. In the proposed architecture, there are two encoder branches which are used to encode the visible and infrared images respectively. Different colours have been used in the architecture to denote different kind of layers which have been elaborately explained in Figure 4. The notation $Layer\char`_i\char`_V[j]$ has been used to better demonstrate the layer structure where 'i' denotes the layer number and j the number of Bottleneck units in the encoder section and Transbasic block in the decoder section. The structure of the Bottleneck unit and the Transbasic block is explained in figure 4. The encoder section extracts several feature maps by virtue of convolution layer. Since we need to obtain a compressed representation of the input data, there is a special block of Bottleneck with downsample operation that helps reduce the dimension of the feature space.
Similarly, we have upsample units in the decoder section(Transbasic Block with upsample operations) to transform reduced feature vector space back to the original dimensions. The upsample and downsample operations are executed by performing convolution and transpose of convolution with a stride of two. The adders in the encoder section at the end of each block are for the fusion of the latent space representations at several levels. The lower half of the network from Layer 5-9 is the decoder section of the autoencoder network, which are all composed of residual layers. The elaborate structure of the layers has been explained in figure 4 part (b) to (g). Initial four layers in the decoder block have upsample units which increase the dimension of feature maps by a factor of two. The residual layers with upsample operation in the decoder section have inverse order in comparison to the residual layers in the encoder section with upsample operation. The CONV and CONV TRANSPOSE blocks shown in figure 4 (e) are the standard PyTorch convolution and convolution transpose operations in which convolution and transpose convolution operators are applied on input image with several input channels. Since the network is deep and we need to overcome the problem of vanishing gradient and provide an effective way of learning simple mapping functions, we have residual connections as denoted by blue lines. The agant layer in the residual connections is composed of 1*1 convolution operator followed by the batch norm layer, which helps to reduce the dimension of feature space which in turn reduces the computation complexity. The discriminator module has the initial layers of network same as the encoder network followed by fully connected layers so as to compile the all the feature maps obtained and the last layer is a sigmoid layer which performs the task of classification and generates a score which is a the probability that a given image is a real image.
\section{Experimentation and Results}
\subsection{Application and Experimental Conditions}

In order to evaluate the performance of the proposed algorithm, we have to use the TNO dataset which contains multi-spectral images, i.e. visual, near-infrared (NIR) and long-wave infrared (LWIR) images for military and surveillance purpose. The images in the dataset are registered spatially, and the corresponding images have pixel-wise correspondence. The training process was carried out on an N-series virtual machine which was equipped with NVIDIA Tesla K80 GPU with NVIDIA GRID 2.0 technology. The fusion process was performed on a Linux based system with 56GB RAM and 340 GB temporary local memory provided by Microsoft Azure platform in python. Multi-modal image fusion has been one of the deeply researched fields in the past due to its extensive practical application in surveillance, military and especially object detection in bad weather conditions since fog, rain and other weather conditions which reduce the visibility in general cause the visible image to not gather the information required and that is where infrared images come in the picture. Firefighters often face difficulty to navigate through smoke filled buildings. However thermal cameras that capture infrared images improves the visibility through thick layer of smoke. Due to the ability of thermal cameras to capture the heat map it can also detect burning surfaces and help find a human in the middle of heavy smoke and flames. Thus image fusion is one of the major problems that should be addressed in recent times. 

\begin{table}
\begin{center}
\begingroup
\setlength{\tabcolsep}{10pt} % Default value: 6pt
\renewcommand{\arraystretch}{1.5} % Default value: 1
\begin{tabular}{ccccccc}
\hline
& VIF        & $Q^{AB/F}$   & SSIM   & MI      & Entropy &        \\ \hline
Athena     & 0.8808 & 0.3066 & 0.7571 & 3.3386 & 7.0596 \\ \hline
Bench      & 2.2271 & 0.5524 & 0.5749  & 3.732   & 7.281  \\ \hline
Bunker     & 2.3969 & 0.2585 & 0.6256  & 3.5013  & 7.0987 \\ \hline
Tank       & 2.3075 & 0.2496 & 0.7470   & 3.9900    & 7.3848 \\ \hline
Sandpath   & 2.0796 & 0.3594 & 0.6478  & 3.1729  & 6.8665 \\ \hline
Nato\_camp & 1.8117 & 0.4077 & 0.70965 & 3.1738  & 6.8165 \\ \hline
Kaptein    & 1.8738 & 0.2423 & 0.6763  & 3.3626  & 7.0055 \\ \hline
Average    & \textbf{1.9396} & \textbf{0.3395} & \textbf{0.6769}  & \textbf{3.4673}  & \textbf{7.0732} \\ \hline
\end{tabular}
\caption{Objective score of proposed fusion method on 7 benchmark image pairs. VIF = visual information fidelity; SSIM = structural similarity; MI = mutual information; EN = entropy}
\end{center}
\endgroup
\end{table}

\begin{table}
\begin{center}
\begingroup
\setlength{\tabcolsep}{10pt} % Default value: 6pt
\renewcommand{\arraystretch}{1.5} % Default value: 1
\begin{tabular}{cccccc}
\hline
         & VIF    & $Q^{AB/F}$ & SSIM   & MI     & EN     \\ \hline
GFF \cite{ref_article30}    & 0.4681 & 0.6180  & 0.4344 & 3.5612 & 6.989  \\ \hline
ASR \cite{ref_article31}      & 0.3767 & 0.5125 & 0.4898 & 2.0770  & 6.4384 \\ \hline
LP \cite{ref_article32}       & 0.4363 & 0.6011 & 0.4938 & 1.9353 & 6.7053 \\ \hline
NSCT \cite{ref_article33}     & 0.4213 & 0.5753 & 0.4945 & 1.883  & 6.585  \\ \hline
SCNN \cite{ref_article19}  & 0.4780  & \textbf{0.6181} & 0.6582 & 2.9402 & \textbf{7.1697} \\ \hline
GTF \cite{ref_article34}      & 0.3440  & 0.3804 & 0.4236 & 2.1623 & 6.5819 \\ \hline
FPDE \cite{ref_article35}    & 0.3338 & 0.4167 & 0.4617 & 1.9024 & 6.3974 \\ \hline
DDCTPCA \cite{ref_article36}  & 0.3927 & 0.5068 & 0.4851 & 1.8382 & 6.5567 \\ \hline
CBF \cite{ref_article37}      & 0.3696 & 0.4752 & 0.4843 & 1.722  & 6.5989 \\ \hline
HMSD \cite{ref_article38}     & 0.3943 & 0.5284 & 0.4891 & 2.6005 & 6.9609 \\ \hline
ADF \cite{ref_article39}      & 0.3270  & 0.3823 & 0.4786 & 2.2094 & 6.3511 \\ \hline
TSIFVS \cite{ref_article40}   & 0.3632 & 0.5059 & 0.4898 & 1.8646 & 6.6270  \\ \hline
Wavelet \cite{ref_article41}  & 0.3028 & 0.2939 & 0.4869 & 2.4895 & 6.3003 \\ \hline
IFEVIP \cite{ref_article42}  & 0.4061 & 0.4805 & 0.4865 & \textbf{3.8723} & 6.8685 \\ \hline
Proposed & \textbf{1.9396} & 0.3395 & \textbf{0.6769} & 3.4673 & 7.0732 \\ \hline
\end{tabular}
\end{center}
\endgroup
\caption{Performance metric evaluation for state of art algorithms}
\end{table}
\subsection{Performance Assessment}
The performance of an image fusion algorithm cannot be judged by objectively evaluating the value of specific performance metrics since the quality of the output depends on the application in which the fused image has to be used. Thus the fusion performance is evaluated on both qualitative and quantitative metrics. Therefore the quality of the fused image can be evaluated using subjective and objective scores. The subjective scores are a function of the visual quality of the fused image and how does the image look perceptually. The perceptual quality depends on how natural the fused image looks visually, the amount of image distortion and the visibility of texture and edge details in the fused image. There is no absolute objective metric that can quantify the quality of the fused image; thus, a combination of several metrics are used to judge the perceptual quality of the fused image. In this paper, we have proposed use of information-based metrics like Entropy and Mutual Information and other parameters which can provide quantitative insight to the image quality like Structural Similarity Index (SSIM), Visual Information Fidelity (VIF) and $Q^{AB/F}$. SSIM is a perceptual quality based model that models image degradation as perceived changes in structural content of the image. Visual Information Fidelity (VIF) is used to compare the quality of the fused image in reference to the input visual and infrared images based on natural scene statistics. It quantifies the amount of information present in the original image and tries to model the amount of information that can be extracted from the fused image which is relevant to its original image. $Q^{AB/F}$ is a gradient-based quality index that measures the amount of edge information in the fused image. The value of all these metrics for different categories of images in the TNO dataset was evaluated and recorded in Table 1.The proposed algorithm was compared to 15 existing state of art algorithms that were surveyed in the most recent paper [43] to compare the fusion performance. It was observed that the values of SSIM and VIF is better than the existing techniques and the value of entropy(EN) and Mutual Information (MI) which are metrics to evaluate the information content in the fused image also exceeds most of the existent algorithms.The results of comparison have been tabulated in Table 2.
\begin{figure}
    \centering
    \includegraphics[height=0.9\textheight]{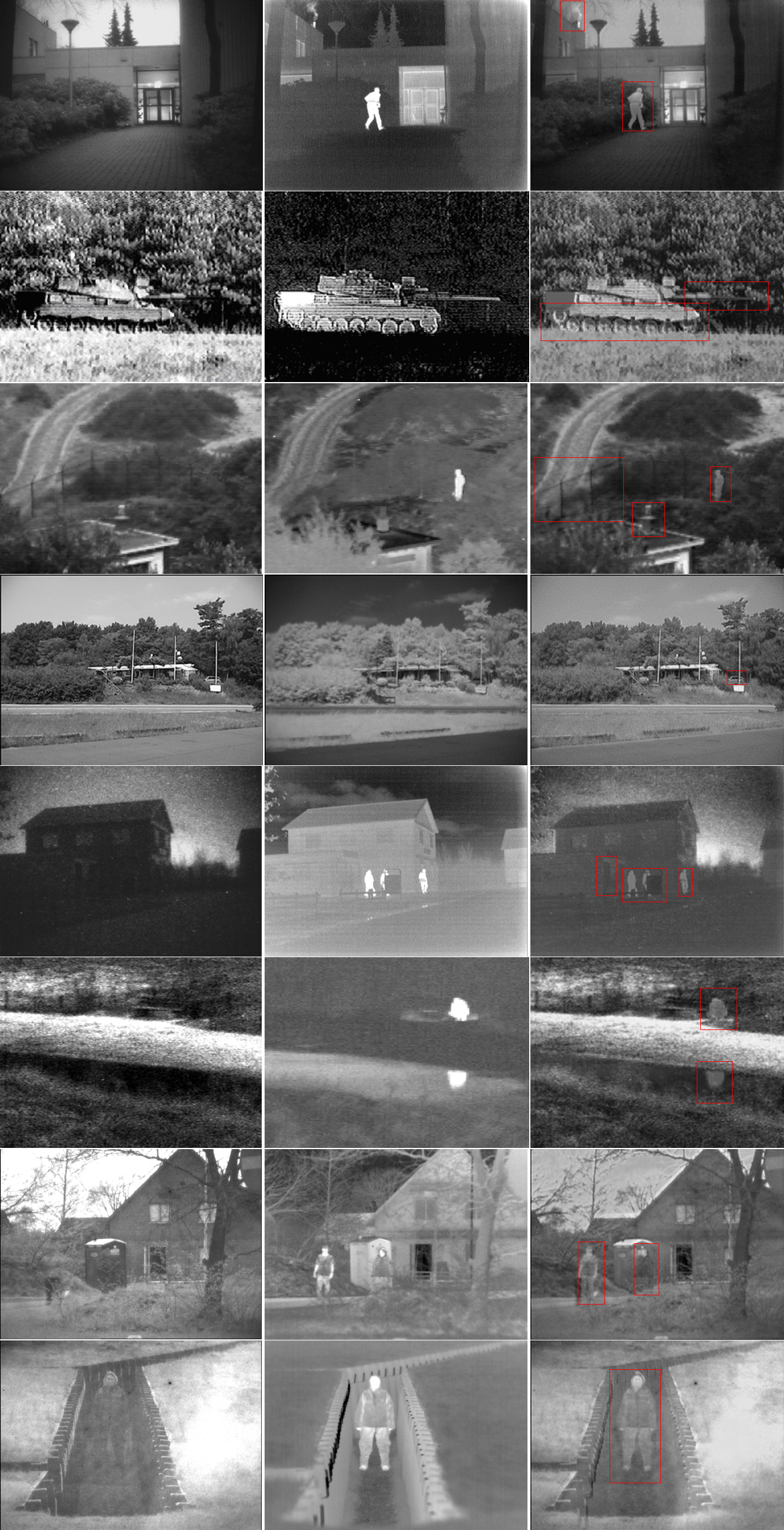}
    \caption{Subjective results of the proposed algorithm on TNO dataset, left to right Visual Image, Infrared image and Fused Image}
\end{figure}
\FloatBarrier
\begin{figure}
    \centering
    \includegraphics[scale=0.3, rotate=-90]{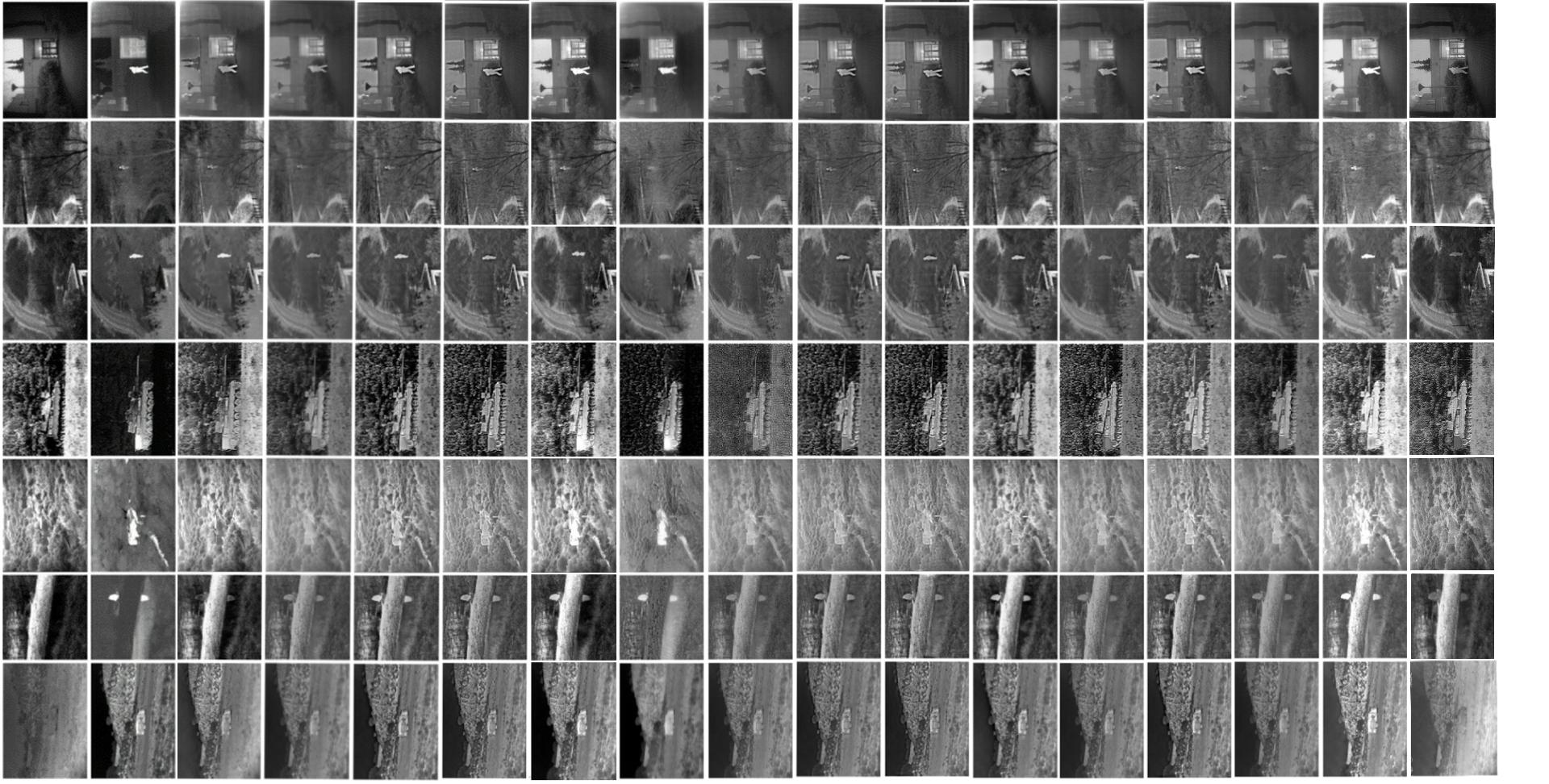}
    \caption{Qualitative metric evaluation results on infrared and visible image pairs from TNO database. From left to right: Athena, Bench, Bunker, Tank, Nato-camp,Sandpath,Kaptein. From top to bottom: Visible,Infrared,GFF, ASR, LP, NSCT, SCNN, GTF,FPDE, DDCTPCA, CBF, HMSD, ADF, TSIFVS, Wavelet, IFEVIP, Proposed}
\end{figure}
\FloatBarrier

\end{document}